# Enhancing Courier Scheduling in Crowdsourced Last-Mile Delivery through Dynamic Shift Extensions: A Deep Reinforcement Learning Approach

Zead Saleh, Ahmad Al Hanbali, and Ahmad Baubaid

*Abstract*— **Crowdsourced delivery platforms face complex scheduling challenges to match couriers and customer orders. We consider two types of crowdsourced couriers, namely, committed and occasional couriers, each with different compensation schemes. Crowdsourced delivery platforms usually schedule committed courier shifts based on predicted demand. Therefore, platforms may devise an "offline" schedule for committed couriers before the planning period. However, due to the unpredictability of demand, there are instances where it becomes necessary to make online adjustments to the offline schedule. In this study, we focus on the problem of dynamically adjusting the offline schedule through shift extensions for committed couriers. This problem is modeled as a sequential decision process. The objective is to maximize platform profit by determining the shift extensions of couriers and the assignments of requests to couriers. To solve the model, a Deep Q-Network (DQN) learning approach is developed. Comparing this model with the baseline policy where no extensions are allowed demonstrates the benefits that platforms can gain from allowing shift extensions in terms of reward, reduced lost order costs, and lost requests. Additionally, sensitivity analysis showed that the total extension compensation increases in a nonlinear manner with the arrival rate of requests, and in a linear manner with the arrival rate of occasional couriers. On the compensation sensitivity, the results showed that the normal scenario exhibited the highest average number of shift extensions and, consequently, the fewest average number of lost requests. These findings serve as evidence of the successful learning of such dynamics by the DQN algorithm.**

## I. INTRODUCTION

The last-mile delivery industry is undergoing significant transformations, with the emergence of innovative models such as crowdsourced delivery, where individuals use their own vehicles to make deliveries for a fee. Crowdsourced couriers login to the delivery platform to indicate their availability while customers place their requests over time. The platform usually does not have full control over the crowdsourced couriers. Thus, this new delivery model poses several difficulties and a host of complex decision-making challenges in terms of planning and execution for delivery platforms.

In this study, we will define two types of couriers: occasional and committed. We refer to couriers as committed if the drivers confirm their availability to the platform to perform deliveries for a particular time period defined by a predetermined start and end time. In contrast, the occasional courier's availability is uncertain and unknown to the platform ahead of time; it begins and ends at the courier's convenience.

In crowdsourced delivery, the platform has to make two main types of decisions: tactical and operational, to manage the delivery capacity such that orders are fulfilled before their due time. These decisions include matching the orders with the couriers and the routing of couriers. Sometimes routing decisions are left to the couriers after knowing their assignment. Moreover, crowdsourced delivery has different decisions compared to traditional delivery models. These include the management of the delivery capacity and its related decisions consisting of online and offline scheduling of committed couriers, as well as deciding the courier compensation. Offline scheduling is concerned with deciding the number of committed couriers needed before starting the service period. Online scheduling, on the other hand, is concerned with adapting committed courier shifts during the service period (e.g., during the day), and can be done on its own (i.e., when no offline schedule is present), or can be used to dynamically adjust offline schedules. Lastly, compensation decisions are concerned with dynamically adjusting the courier payments to increase the probability that they accept an assigned request, and adjusting compensation may also increase the probability of occasional couriers entering the system. In this research, we address the dynamic and stochastic nature of last-mile delivery with a particular focus on online scheduling decisions within the context of crowdsourced delivery.

In the literature, previous research in crowdsourced delivery has predominantly concentrated on routing and order assignment decisions, leaving online and offline courier scheduling decisions underexplored. This work focuses on a

This work was supported by the Interdisciplinary Research Center of Smart Mobility and Logistics, King Fahd University of Petroleum and Minerals, Dhahran 31261, Saudi Arabia. *(Corresponding author: Zead Saleh)*
Zead Saleh, Ahmad Al Hanbali, and Ahmad Baubaid are with the Industrial and Systems Engineering Department, King Fahd University of Petroleum and Minerals, Dhahran 31261, Saudi Arabia. E-mail: g201080800@kfupm.edu.sa; ahmad.alhanbali@kfupm.edu.sa; baubaid@kfupm.edu.sa.

scenario where a delivery platform must dynamically adjust its offline schedule for committed couriers over the course of a day while accommodating occasional couriers who can fulfill delivery requests. We consider a simple adjustment mechanism whereby committed couriers are notified and are asked if they are willing to extend their shifts for a higher compensation when the platform anticipates the need for more committed couriers. To the best of our knowledge, we are the first to study committed crowdsourced couriers' shift extension as an online decision in last-mile delivery settings. The primary objective is to maximize the platform's profit over the course of the day by efficiently matching couriers with customer requests. Requests arrive randomly over the day with specific attributes such as pickup and delivery locations, time constraints, and associated costs. The platform faces penalties for lost orders and must also consider compensation for both committed and occasional couriers.

We propose a sequential decision process framework that models this problem. This framework allows us to formalize the scheduling decisions and capture the interactions between couriers, requests, and the platform's operations. Our objective is to find a policy that maximizes the expected profit of the platform by making informed decisions regarding shift extensions and request assignments.

Traditional approaches to scheduling in crowdsourced delivery have primarily relied on heuristics and optimization algorithms, see, e.g., [1]. In our research, we explore the application of Deep Reinforcement Learning (DRL) to address online scheduling challenges in crowdsourced last-mile delivery. DRL is a machine learning paradigm designed for sequential decision-making problems, making it a promising tool for handling the dynamic and stochastic nature of scheduling decisions in this context. By exploiting the deep reinforcement learning technique in the scheduling challenges of crowdsourced last-mile delivery, we aim to contribute to the advancement of more efficient and responsive delivery operations in the home delivery business industry. The main contributions of this research can be summarized as follows:

• Studying the benefits of making online adjustments to offline schedules for committed couriers in crowdsourced delivery by dynamically extending couriers' shifts.

• Developing a sequential decision process model for the problem of extending committed courier shifts.

• Proposing a Deep Q-Learning (DQN) solution approach for our proposed shift extension model.

• Evaluating the effectiveness of the DQN solution approach by comparing the performance of the resulting policy with that of a baseline policy where shift extensions are not an option.

• Analyzing the sensitivity of the learned policy's performance by systematically varying specific input parameters of the model and examining their effects on key output metrics.

The subsequent sections of this paper are organized as follows. Section II presents a brief survey of the related literature. In Section III, the problem is defined and formulated as a sequential decision process. Section IV introduces the DQN solution approach and outlines the methodology and techniques employed to address our problem. Section V presents empirical evidence that demonstrates the effectiveness of the proposed solution approach, and a sensitivity analysis study to gauge the flexibility and robustness of the proposed policy. Finally, in Section VI, the work is summarized, highlighting the key findings and contributions, and future research directions are suggested.

## II. RELATED WORK

In the context of crowdsourced delivery, various decisions need to be made, including assigning orders to couriers, courier routing, courier online and offline scheduling, and determining couriers' compensation. While routing and order assignment decisions have received significant attention in the literature, only a few studies have focused on online and offline scheduling decisions. In offline scheduling, shifts are created and made available to crowdsourced couriers before the start of the planning horizon. Couriers who choose to work during the specified shifts are considered scheduled (committed) couriers, while unscheduled (occasional) couriers may arrive at any time according to their preference during the operational period to fulfill orders. In this section, we discuss the related literature, starting with courier scheduling decisions in crowdsourced delivery. We then discuss the application of reinforcement learning techniques in crowdsourced last-mile delivery.

### A. Courier scheduling decisions in crowdsourced delivery

For offline scheduling, [2] proposed a method that determines full-time shifts for fully committed couriers. [3] developed an optimization model to determine the optimal workforce size. [4] proposed an offline scheduling model for fully committed couriers, specifying the number of shifts, their starting times, and durations. [5] modified the work of [4] and applied a machine learning approach to determine the number of fully committed couriers and their shifts. For online scheduling, [1] investigated a proactive approach where couriers are requested through notifications to commit to a specific time period (block) in advance, taking into account the expected future demand. The authors also considered factors such as the probability that couriers will accept, and the timing of at which these notifications are sent. Based on the previous literature review, there is a limited number of studies focusing on online scheduling with the consideration of uncertainty in acceptance of delivery tasks and routing.

### B. Reinforcement learning for crowdsourced last-mile delivery

In this section, we will focus on reviewing the literature that utilizes reinforcement learning (RL) in the context of urban logistics with crowdsourcing. RL is a machine learning paradigm designed for addressing sequential decision-making problems through a process of exploration and exploitation within an environment. In RL, an entity, called agent, interacts with the environment to learn the best decision based on the received reward. The agent assesses the environment's current state and makes decisions based on a function that maps a state to an action, called a policy. The primary objective of the agent is to learn the optimal policy that maximizes the value function defined as the cumulative discounted reward it expects to receive by deciding on the best actions.

While there has been significant research on various aspects of courier decisions, including routing, acceptance, and offline scheduling, there is a lack of focus on online scheduling, specifically on the application of RL techniques to address the critical scheduling decisions. In particular, RL, with its ability to handle sequential decision-making problems, especially those with a large state space, could be a valuable tool for optimizing online schedules where shifts are dynamically adjusted based on real-time demand, courier availability, and offline schedules. Due to the extremely large size of the state space in this problem, it is not feasible to enumerate all possible states. Additionally, using tabular RL methods like tabular Q-learning (QL) to approximate the value function is impractical. In contrast, the use of Deep Q-Networks (DQNs) may help overcome these practical challenges. By leveraging neural networks, DQNs learn the relationships between different states and actions. This learned information can then be used to extrapolate values from explored state-action pairs to unexplored pairs, alleviating the memory and exploration limitations associated with tabular methods.

Regarding the routing decision, [6] addresses the pickup and delivery problem (PDP), a variant of the vehicle routing problem with the constraint that each pickup node must precede its paired delivery node. The authors propose a deep reinforcement learning method with a heterogeneous attention mechanism to help the policy network learn precedence relations and roles of different nodes, which experimental results show achieves better performance than existing methods on larger problems and new distributions. Another research aims to optimize an on-demand delivery problem by formulating it as a generalized scheduling model to address its dynamic complexities [7]. A novel reinforcement learning based framework is introduced where an algorithm is trained offline using expert demonstrations and then utilizes the learned model for online decision making, demonstrating improved customer satisfaction and efficiency.

In the context of crowdsourced delivery problems, [8] introduced a DQN algorithm geared towards multi-hop crowdsourced delivery using public transport, factoring in the conditions of public transport systems. Their DQN model acquires knowledge about both parcels and passengers to make dispatching decisions. This approach differs from previous research on multi-hop delivery, which primarily focused on acceptance and assignment. [9] proposed a DQN approach to optimize same-day delivery using vehicles and drones. Through computational analysis, their policy outperforms benchmark policies, demonstrating the effectiveness of their approach in dynamically assigning customers to the appropriate delivery method. Numerous studies have combined heuristics and RL methods. For instance, [10] trained a DQN agent to select heuristics, such as node exchange and insertion within and between carriers, based on delivery instructions. They also devised a Tabu list and a priority list to prevent agents from repeatedly generating the same sequences and to promote exploration. In another context, [11] introduced a DQN approach to determine whether to accept pickup and delivery orders using a single vehicle. Their framework considers parcel location, weight, vehicle information (location, schedule, battery status, and capacity), and makes acceptance decisions accordingly. This method employs only one vehicle, eliminating the need for order assignment decisions. In a subsequent paper, [12] extended this approach to handle multiple vehicles. [13] explored a similar problem, where drivers can reject orders and reassignment can occur within their Q-learning simulation model. Recently, [14] addressed the problem of dynamically adding couriers to meet delivery demands. By applying a DQN algorithm, they developed a policy that balances the cost of adding couriers with the quality of service. Similar to our work, the latter study focuses on online scheduling decisions. However, there are notable differences between our study and theirs. Firstly, they developed a learning policy to add on-call couriers as needed, while our policy learns the best shift extensions of committed couriers who are already working. In their model, they assumed that on-call couriers are under full control of the platform and are available whenever needed, whereas our model incorporates uncertainty regarding the acceptance of shift extensions by the (committed) couriers. Secondly, we consider two types of couriers with different commitment characteristics and compensation schemes, namely committed and occasional, whereas they only considered committed couriers who are on shift or on call. Occasional couriers can help in increasing the number of requests served at a competitive price as they are, typically, cheaper than committed couriers. This will be demonstrated through the sensitivity analysis in Section V; specifically, as we increase the number of occasional couriers, the total shift extension cost paid to committed couriers throughout the planning horizon reduced significantly. Thirdly, they used a simple heuristic for assigning orders to couriers whereas we consider the assignment decision as part of the action space (with corresponding costs) at each decision epoch in addition to the shift extensions.

4Our literature review has identified very limited research that specifically incorporates RL into online or offline scheduling decision-making processes for crowdsourced delivery. Therefore, the research gap lies in the absence of studies that explore how RL, namely deep Q-learning, can be effectively applied to optimize these critical scheduling decisions, considering factors like courier availability, order demand, and operational efficiency.

## III. MODEL FORMULATION

The following section describes the mathematical model and notation used to formally define the sequential decision process for the considered dynamic scheduling problem, including details on the states, actions, and stochastic components over multiple time periods.

### A. Problem Definition

We consider the problem of dynamically adjusting the offline schedule of committed couriers (using shift extensions) while accounting for occasional couriers that is faced by a pickup and delivery platform over a time horizon $\mathcal{H} = [0, H]$, typically corresponding to a day. Requests arrive dynamically throughout the operating period $\mathcal{H}$ and can be fulfilled by either committed or occasional couriers. Let $\mathcal{P}$ denote the set of all possible pick-up locations, and $\mathcal{D}$ denote the set of all possible delivery locations. The platform aims to match couriers with customer requests. Each request, $r$, consists of a request arrival time $\tau_r^{req} \in \mathcal{H}$, a pickup location $p_r \in \mathcal{P}$, a delivery location $d_r \in \mathcal{D}$, and a latest feasible assignment time, $l_r^{req} \in \mathcal{H}$. Thus, each request is represented by aforementioned attributes as $(\tau_r^{req}, l_r^{req}, p_r, d_r)$. The platform aims to match couriers with customer requests. For simplicity, we consider a request fulfilled as soon as it is assigned to a courier, i.e., we assume that couriers will accept and perform all requests assigned to them. A request should be assigned to a courier before $l_r^{req}$ after which it is considered lost. The platform incurs a per-request penalty cost, $\theta$, for each lost request. Lost requests are captured by a binary variable, $\beta_r$, that is equal to 1 if $a_r \notin [\tau_r^{req}, l_r^{req}]$, where $a_r$ is the request assignment time, and 0 otherwise. For each assigned request, the platform collects a revenue of $f_r$.

Couriers can be either committed or occasional. Each committed courier is defined by a shift starting time, $e_m^{com} \in \mathcal{H}$, a shift ending time, $l_m^{com} \in \mathcal{H}$, and a location at time $t$, $o_m^{com}(t)$. All the committed couriers will initially appear at one of the pickup locations. Moreover, to indicate whether a committed courier is busy at time $t$, $e_m^{com} \leq t \leq l_m^{com}$, we set a binary variable $\varphi_m$ to 1, and 0 otherwise. Therefore, a committed courier, $m$, is defined by the aforementioned attributes, namely $(e_m^{com}, l_m^{com}, o_m^{com}(t), \varphi_m)$. Each occasional courier, $n$, is identified by his/her entering time, $\tau_n^{occ} \in \mathcal{H}$, and location $o_n^{occ}$. Thus, each occasional courier is denoted by the mentioned attributes as $(\tau_n^{occ}, o_n^{occ})$. Moreover, if the platform assigns a request to an occasional courier we remove her/him from the list of available occasional couriers. This is without loss of generality since if the occasional courier wants to stay and deliver more requests, they can be considered as a new occasional courier entering the system. The role of the platform is to properly adjust the existing offline schedule and to match requests and couriers in a way that maximizes the platform's profit. Specifically, the goal is to assign incoming requests to available couriers in a way that maximizes the total profit over the time horizon.

We discretize our time horizon into equidistant decision epochs $t \in \{1, ..., T\} \subset \mathcal{H}$. The time between any two successive decision epochs is called a *period* and has a fixed length $\sigma$. Periodically, at each decision epoch, $t$, the platform revises the offline schedule and offers the committed couriers whose shifts end at the next decision epoch to extend their shifts for the following $\eta$ decision epochs for higher compensation; this is done through a notification system. For example, if we are now at decision epoch $t = 3$, all couriers who finish their shifts by $t = 4$ will receive a notification asking them to extend their shift for $\eta = 2$ periods. In this case, the couriers that will accept the shift extension will become available until epoch $t = 6$. If a courier decided to extend her/his shift in decision epoch $t$, she/he will receive a compensation of $\rho$ per *period*.

Together with the penalty for lost requests, the compensation paid to the committed and occasional couriers make up the platform's costs. Couriers are paid by the platform differently according to their type. Specifically, committed couriers are paid a per-period wage, $\omega^c$, for each *period* of their shifts regardless of their utilization. Additionally, they receive a per-unit distance amount, $\nu^c$, for their entire route: from the starting position, through the pickup location, and ending with the destination. On the other hand, occasional couriers are paid a fixed amount per request fulfilled, $\omega^o$, and a per-unit distance payment, $\nu^o$, for the distance between the pickup and destination locations. Note that unlike committed couriers, occasional couriers are not compensated for the initial trip segment from their starting location to the pickup location. Moreover, the platform does not know where and when the occasional couriers will enter and leave the system.

### B. Sequential decision process model

This section formally defines the mathematical model and its notation. We describe the sequential decision process components following Powell's framework [15]. Notably, our model has two types of decisions: sending shift extension notifications and assigning requests to couriers.

- **Preparation**

The set of couriers changes dynamically as new occasional couriers enter the system, some committed couriers start their shifts, and other couriers leave for good. Similarly, for the set of requests, new requests enter, and lost and assigned orders leave the system. To keep track of all committed couriers, we assume that, at each decision epoch $t$, we have a set of committed couriers $\boldsymbol{M}_t = \{m_1, ..., m_{c_t}\}$ with $c_t$ denoting the overall number of committed couriers observed up to t. We denote the subset of committed couriers who are currently on shift at decision epoch t as $\boldsymbol{M}_t^s = \{m \in \boldsymbol{M}_t \mid e_m^{com} \leq t \leq l_m^{com}\}$. For ease of presentation, we do the same for the number of overall requests at time $t$, $\boldsymbol{R}_t = \{r_1, \ldots, r_{i_t}\}$ with $i_t$ representing the overall number of requests observed up to $t$.

- **States**

The state at decision epoch $t$, denoted by $S_t$, represents the information and features of the system that are considered important for the decision-making. Specifically, $S_t$ is constructed using three main components: the set of present requests, $\boldsymbol{R}_t^a = \{r \in \boldsymbol{R}_t | \tau_r^{req} \leq t \leq l_r^{req}, \beta_r = 0\}$, the set of available committed couriers, $\boldsymbol{M}_t^a = \{m \in \boldsymbol{M}_t | e_m^{com} \leq t \leq l_m^{com}, \varphi_m = 0\}$, and $\boldsymbol{N}_t^a$ the set of present occasional couriers. Each of these components has its associated attributes and parameters. More precisely, the set of present requests, $\boldsymbol{R}_t^a$, includes individual requests with their associated attributes as described in Section III-A The set of available committed couriers, $\boldsymbol{M}_t^a$, comprises committed couriers who are currently free and ready for assignment at epoch $t$. Each committed courier $m$ is characterized by his associated attributes as described in Section III-A. Similarly, the set of present occasional couriers, $\boldsymbol{N}_t^a$, represents couriers who are available for assignments at epoch $t$. Each occasional courier $n$ is defined by it associated attributes as described previously. In summary, the state at decision epoch $t$ is given by $S_t = (\boldsymbol{R}_t^a, \boldsymbol{M}_t^a, \boldsymbol{N}_t^a)$. In the following, we denote by $\boldsymbol{K}_t^a$ the union of $\boldsymbol{M}_t^a$ and $\boldsymbol{N}_t^a$.

- **Action**

At every decision epoch, $t$, an action $a_t$ is chosen from a pool of potential actions $X(S_t)$. An action incorporates the following types of decisions: (1) which committed courier(s) to notify for extending their shifts, and (2) which request is assigned to which courier.

We represent an action at epoch $t$ as $a_t = (x_{mt}, z_{rkt})_{m \in \boldsymbol{M}_t^s, t \in \mathcal{H}, r \in \boldsymbol{R}_t^a, k \in \boldsymbol{K}_t^a}$, where $x_{mt}$ represents the extension decision variable and $z_{rkt}$ represents the assignment decision variable. We define each of these variables in turn. The extension decision is modelled as a binary variable, $x_{mt}$, which is equal to 1 if courier $m \in \boldsymbol{M}_t^s$ is notified to extend his shift at epoch $t$, and 0 otherwise. The notification is sent to the courier one epoch ahead. For example, the notification is sent at epoch $t$ for the couriers ending their shifts at epoch $t+1$ to extend their shifts for an additional $\eta$ *periods*. An action is considered feasible if it satisfies the following constraint:

$$\sum_{\substack{m \in \boldsymbol{M}_t^s: \\ l_m^{com} = t+1}} x_{mt} \leq \alpha_t \quad (1)$$

This constraint ensures that, at decision epoch $t$, we only consider for extension committed couriers whose shifts end in epoch $t+1$ and it stipulates that the number of notified couriers does not exceed a threshold $\alpha_t$ that is specified by the platform. The second part of the action corresponds to the assignment decision, and is represented by $z_{rkt}$, a binary variable equal to 1 if order $r \in \boldsymbol{R}_t^a$ is assigned to courier $k \in \boldsymbol{K}_t^a$ at decision epoch $t$, and 0 otherwise. The assignment decision is considered valid if every courier is assigned at most one request and every request is assigned to at most one courier. This is described mathematically by constraints (2) and (3) below, respectively.

$$\sum_{r \in \boldsymbol{R}_t^a} z_{rkt} \leq 1, \quad \forall k \in \boldsymbol{K}_t^a \quad (2)$$

$$\sum_{k \in \boldsymbol{K}_t^a} z_{rkt} \leq 1, \quad \forall r \in \boldsymbol{R}_t^a \quad (3)$$

- **Stochastic Information and Transition**

After notifying the candidate committed couriers, the platform observes their instantaneous responses to the extension notification. This is modeled as binary parameter $y_{mt}$ where a value of 1 means courier $m$ accepts to extend her/his shift at epoch $t$, and 0 otherwise. Note, if a courier accepts the notification the value of $y_{mt}$ will be kept equal to one while the driver is in her/his shift extension period. The response of the couriers to the notification follows a Bernoulli distribution with success probability $pr_m$. Then, we update the available committed courier set $\boldsymbol{M}_t^a$ and modify the corresponding attributes according to this response, e.g., the shift end time, $l_m^{com}$, of each courier accepting the shift extension is adjusted based on their response. The new state at decision epoch $t+1$, $S_{t+1}$, is a combination of how every element of the state $S_t$ variable changes given the action $a_t$, as well as stochastic information on the request arrivals and courier extension acceptance. The new sets $\boldsymbol{M}_{t+1}^a$, $\boldsymbol{N}_{t+1}^a$ and $\boldsymbol{R}_{t+1}^a$ are updated as follows:

a) $\boldsymbol{M}_{t+1}^a = (\boldsymbol{M}_t^a \cup \boldsymbol{M}_t^+ \cup \boldsymbol{M}_t^{++}) \setminus \boldsymbol{M}_t^-$ (4)

a) $\boldsymbol{N}_{t+1}^a = (\boldsymbol{N}_t^a \cup \boldsymbol{N}_t^+) \setminus \boldsymbol{N}_t^-$ (5)

b) $\boldsymbol{R}_{t+1}^a = (\boldsymbol{R}_t^a \cup \boldsymbol{R}_t^+) \setminus (\boldsymbol{R}_t^- \cup \boldsymbol{R}_t^{--})$ (6)

where:

$\boldsymbol{M}_t^+$: The set of new committed couriers starting their shifts in period $(t, t+1]$. This is known in advance from the offline schedule.

$\boldsymbol{M}_t^{++}$: The set of committed couriers that accepted to extend their shifts at epoch $t$ (which depends on the acceptance response vector **y**).

$\boldsymbol{M}_t^-$: The set of committed couriers who finish their shifts in period $(t, t+1]$. This is known in advance from the offline schedule.

$\boldsymbol{N}_t^+$: The set of new occasional couriers that join the system in period $(t, t+1]$. This is stochastic information, where the occasional couriers join the platform over time following a stochastic process, e.g., a Poisson process.

$\boldsymbol{N}_t^-$: The set of occasional couriers that have left the system in period $(t, t+1]$. A courier leaves the system if she/he is assigned a request or she/he has been waiting for a specific time (modeled as a random variable), $\psi$, without an assignment.

$\boldsymbol{R}_t^+$: The set of new requests that have arrived in period $(t, t+1]$ where the new requests arrive to the system according to a stochastic process, e.g., a Poisson process.

$\boldsymbol{R}_t^-$: The set of assigned or served requests in period $t$ (which depends on decision variable **z**).

$R_t^{--}$: The set of lost requests in period $(t, t+1]$. This set captures requests who past their maximum assignment time, $l_r^{req}$, without being assigned to any couriers.

We also update the status of the assigned committed couriers and set their earliest free time to be equal to the expected delivery time, $\mu_r$, of the request that has been assigned to them. We use the Euclidian distance between pickup and delivery locations and average speed of $\lambda$ to calculate $\mu_r$. Furthermore, the binary variable, $\beta_r$, that captures the lost requests is also updated based on the assignment decision variable, $z_{rkt}$.

- **Rewards**

The reward function $R(S_t, a_t)$ calculates the overall reward obtained from a given a state and action and is given by:

$$R(S_t, a_t) = \sum_{k \in K_t^a} \sum_{r \in R_t^a} f_r z_{rkt} - \omega^c NC_t^s - \nu^c \sum_{k \in M_t^a} \sum_{r \in R_t^a} (\delta_{o_m^{com}(t), p_r} + \delta_{p_r, d_r}) z_{rkt} - \omega^o \sum_{k \in N_t^a} \sum_{r \in R_t^a} z_{rkt} - \nu^o \sum_{k \in N_t^a} \sum_{r \in R_t^a} \delta_{p_r, d_r} z_{rkt} - \theta \sum_{r \in R_t^a} \beta_{rt} - \rho \sum_{m \in M_t^a} y_{mt}. \quad (7)$$

In Equation (7), the first term represents the revenue generated by the platform for assigned/served requests. The second term accounts for the per-*period* wage provided to committed couriers where $NC_t^s$ represents the number of committed couriers working (on shift) at decision epoch $t$. The third term captures the per-unit distance compensation for committed couriers for the trip from their location to the request's pickup location, $\delta_{o_m^{com}(t), p_r}$, and then from there to the delivery point, $\delta_{p_r, d_r}$. The fourth and fifth terms represent the compensation for occasional couriers per order and per unit distance, respectively. Note that the per-unit distance compensation is only incurred for the distance between the request's pickup and delivery points, $\delta_{p_r, d_r}$. The sixth term represents the penalty for lost requests. Lastly, the seventh term represents additional compensation paid to committed couriers who agree to extend their shifts.

- **Objective function**

A solution to the problem is a decision policy $\pi$ which maps each state $S_t$ to an action $a_t$. The objective is to find a policy $\pi^*$ maximizing the expected discounted total reward of the platform:

$$\pi^* = \arg\max_{\pi \in \Pi} \mathbb{E}\left[\sum_{t=0}^{T} \gamma R(S_t, X_t^\pi(S_t)) \mid S_0\right] \quad (8),$$

where $X_t^\pi(S_t)$ is the action selected by policy $\pi$, $\gamma$ is the discount factor, $R(S_t, X_t^\pi(S_t))$ is the reward function defined in equation (7), and $S_0$ is the initial state. In the following section, we shall detail the solution approach to solve the sequential decision process model presented above.

## IV. DEEP REINFORCEMENT LEARNING FOR STOCHASTIC LAST-MILE DELIVERY WITH CROWDSOURCING

Reinforcement learning (RL) belongs to a set of machine learning techniques, alongside supervised learning and unsupervised learning [16]. RL focuses on training an agent to optimize its actions by accumulating and learning from experiences gained through interaction with the environment. The objective is to find the best policy $\pi^*$ that maximizes the total expected discounted reward over a specific time frame by taking sequential actions. At each decision point, the agent possesses information about the current state of the environment and selects the optimal action based on its experience. This action causes the environment to transition to a new state, and the agent receives a reward, or reinforcement, as an indication of the action's quality. RL is particularly suited for scenarios involving sequential decision processes, where the optimal action depends on the current state and expected future states.

In this section, we introduce our solution approach, noting that the Bellman optimality equation (9) is a well-known theoretical means of finding a solution to a sequential decision process model:

$$V^*(S_t) = \max_{a_t \in X(S_t)} \{R(S_t, a_t) + \gamma \, \mathbb{E}[V^*(S_{t+1}) \mid S_t]\} \quad (9),$$

where $V^*(S_t)$ is the value function of state $S_t$, $R(S_t, a_t)$ is the immediate reward for taking action $a_t$ (extending shifts and assigning orders decisions in our setting) when in state $S_t$, $\gamma$ is the discount factor that determines the relative importance of immediate rewards compared to future rewards in a sequential decision-making problem (a value of $\gamma$ closer to 1 indicates a greater emphasis on long-term rewards, while a value closer to 0 places more importance on immediate rewards), and $S_{t+1}$ represents the next state after taking $a_t$ action in state $S_t$.

The Deep Q-Network (DQN) is an influential reinforcement learning technique that merges deep neural networks with Q-learning. It was first introduced by [16] and is widely applied to address sequential decision problems across various domains. This section outlines how we have adapted the DQN, our chosen training algorithm, to the context of our problem.

In RL, the training and testing phases are key stages in the learning process of an RL agent. In the training phase, the RL agent interacts with the environment to learn and improve its decision-making abilities. The agent explores the environment by taking actions based on its current policy and receives feedback in the form of rewards from the environment. It uses this feedback to update its policy function, aiming to maximize the cumulative discounted reward over time. Once the RL agent has been trained, it proceeds to the testing phase. In this phase, the agent's learned policy is then fixed (i.e., it will not be updated any further) and evaluated on held-out data to assess its performance and how well the policy generalizes.

In the DQN training process, the agent undergoes multiple episodes (multiple decision epochs, e.g., an operating day), with each episode focusing on enhancing the solution incrementally through a sequence of actions, as described in Section III-B. These actions are taken by the agent step-by-step within each episode. After selecting an action, the agent observes a reward and the environment transitions to a new state.

At every decision epoch, the agent employs an ε-greedy strategy, a balance between exploration and exploitation. Exploration occurs with probability $\varepsilon$ (chosen a priori) and entails selecting an action randomly, while exploitation occurs with probability $1 - \varepsilon$, and involves choosing the best-known action type based on the agent's learned experience using the so-called Q-value of a state-action pair which represents the expected future reward the agent will receive if it takes action $a_t$ in state $S_t$ and follows the optimal policy from that point onward. Balancing exploration and exploitation is crucial. Initially, when the agent lacks learned experience, $\varepsilon$ is set to be very close to 1 to emphasize exploration. As time progresses, $\varepsilon$ gradually decreases until reaching a minimum value called *Min ε*, favoring the exploitation of learned actions. This decay rate, denoted by $\xi$, determines the shift in probabilities between consecutive decision epochs (for the whole training period) where $\xi$ serves as a hyperparameter. At each epoch $t$, $\varepsilon$ updated as follows $\varepsilon_t = \varepsilon_{t-1}(1 - \xi)$.

DQN features an *experience replay*, which involves utilizing a replay memory, RM, to store the agent's experiences during training, with a maximum capacity of experiences. Each experience is associated with taking an action at a specific state and decision epoch, observing state transitions, and obtaining rewards. Initially, the replay memory is empty. As training progresses, experiences accumulate and are added to RM. When the capacity is reached, adding a new experience requires removing the oldest one.

At each decision epoch, a Deep Neural Network (DNN) is trained using a minibatch $M_{sub}$, consisting of randomly selected samples from RM. Initially, when the number of accumulated experiences in RM is less than $|M_{sub}|$, experiences continue to accumulate in RM, but updating the weights in training is deferred until $|M_{sub}|$ experiences are available. Using randomly selected minibatch samples for experience replay offers several advantages: it reduces sample correlations, enhances learning efficiency, and allows each experience to contribute to multiple weight updates, improving data efficiency. Additionally, experience replay averages the behavior distribution across previous states, promoting smoother learning and preventing parameter oscillation or divergence.

In the realm of DQNs, two key components are the *Q-network* (action-value network) and the *target network*. The Q-network is a deep neural network that approximates Q-values denoted by $Q(s_t, a_t, w)$, for various actions in a given state, where Q-values represent the expected future discounted rewards for taking specific actions in a state and $w$ refers to the weights of the neural network used in the DQN.

Conversely, the target network, a duplicate of the Q-network, is employed to estimate target Q-values during training. The target network is initially initialized with the same parameters as the original network. Rather than updating the target network parameter ($\widetilde{w}$) with those of the original network ($w$) at each decision epoch, the target network's parameters remain frozen for a specified number of decision epochs, denoted by $\zeta$. Only after $\zeta$ decision epochs will the target network parameters be updated to match the current values of the original network parameters. This approach provides a stable estimation of Q-values, and enhances the convergence of the DQN algorithm. Ideally, the optimal target Q-value should satisfy the Bellman optimality equation (9), which can be expressed as:

$$Q^*(S_t) = R(S_t, a_t) + \gamma \max_{a_{t+1}} Q(S_{t+1}, a_{t+1}: \widetilde{w}) \quad (10)$$

where $\max_{a_{t+1}} Q(S_{t+1}, a_{t+1}: \widetilde{w})$ is the maximum Q-value predicted by the target network for each next state action pair. During training, the Q-network is updated using backpropagation and the Adam optimizer, a popular gradient descent-based algorithm for minimizing loss. The Q-network adjusts its weights and biases to minimize the difference between predicted Q-values and target Q-values. Typically, DQN employs the mean squared error (MSE) loss function to quantify the disparity between predicted and target Q-values:

$$L(w) = \frac{1}{|M_{sub}|} \sum_{(S_t, a_t, R(S_t, a_t), S_{t+1}) \in M_{sub}}^{|M_{sub}|} \left[ R(S_t, a_t) + \gamma \max_{a_{t+1}} Q(S_{t+1}, a_{t+1}: \widetilde{w}) - Q(S_t, a_t: w) \right]^2 \quad (11)$$

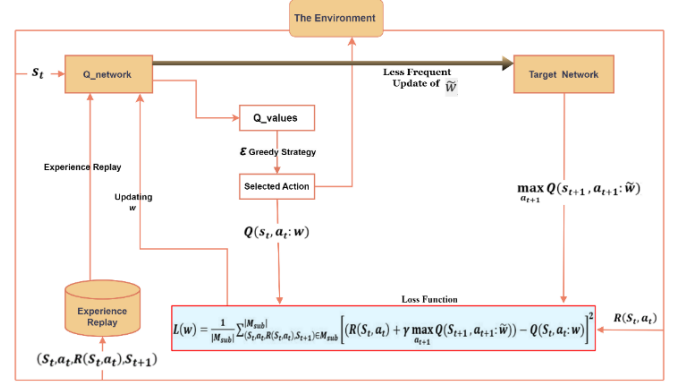

**Fig. 1.** The architecture of the DQN algorithm

The agent continually interacts with the environment, collecting experiences and iteratively updating the Q-network. Once an episode (one day of operations) is finished, we restart the process from the same initial state. Over time, the Q-network becomes more adept at approximating optimal Q-values for each state-action pair, allowing the agent to make informed decisions and accumulate higher cumulative rewards. Figure 1 provides an overview of the DQN algorithm's steps.

V. NUMERICAL EXPERIMENTS

In this section, we illustrate the practical implementation of our solution approach described in Section IV. Section V-A represents the general setup of the problem including parameter settings. In Section V-B, we gauge the effectiveness of the training phase by analyzing the model's performance with respect to metrics such as average total reward and loss over episodes. Next, in Section V-C, we subject the proposed shift extension policy to rigorous benchmarking. This entails a comparison with a baseline policy where there are no shift extensions. We evaluate both policies in terms of reward and requests fulfilled. Finally, in Section V-D, we perform sensitivity analysis to check the robustness of the proposed crowdsourced delivery model against changes in input parameters, and obtain more insights about the dynamics of the problem. The DQN algorithm is implemented and trained using the PyTorch framework, and all numerical experiments are executed on a computer equipped with an Intel(R) Xeon(R) W_1290P processor with a clock speed of 3.70GHz, and 32GB of RAM.

*A. Setup*

As we assume a given offline schedule for committed couriers, we start our experiments by constructing an offline schedule that is designed with the presence of occasional couriers in mind. In our experimental setup, we designate a total of 50 committed couriers scheduled throughout the day's duration, while occasional courier and request arrivals follow Poisson distributions with means 1 and 2 per period (time between any consecutive decision epochs), respectively. The requests' pickup points are randomly selected from a pool of 40 distinct points, and the same applies to the delivery points. Figure 2 shows a map of the distribution of pickup and delivery points. As for the starting points of both occasional and committed couriers, they are randomly located on the map.

As mentioned in Section III-B, our state space is dynamic, evolving at each decision epoch in accordance with the current number of committed couriers, occasional couriers, requests, and their associated attributes. To prepare the state vector for input into the DQN, the relevant features that describe the current state of the environment were normalized using a min-max scaler. This process, as suggested by [17], enhances the accuracy of relative importance among the considered features. Detailed cost parameters, expressed in monetary unit (MU), considered in our reward function are summarized in Table II, while time related inputs are summarized in Table III. For the DQN, each episode in our experiments consists of 200 decision epochs (the time between the decision epochs is 6 minutes), effectively covering a full day's operation spanning 20 hours. At the beginning of each episode, the initial state, $S_0$, is configured with four available requests, two committed couriers, and three occasional couriers. Through preliminary experiments, we were able to determine an effective neural network architecture for the Deep Q-Network (DQN), namely a configuration of four hidden layers, each composed of 64 neurons utilizing Rectified Linear Unit (ReLU) activation functions.

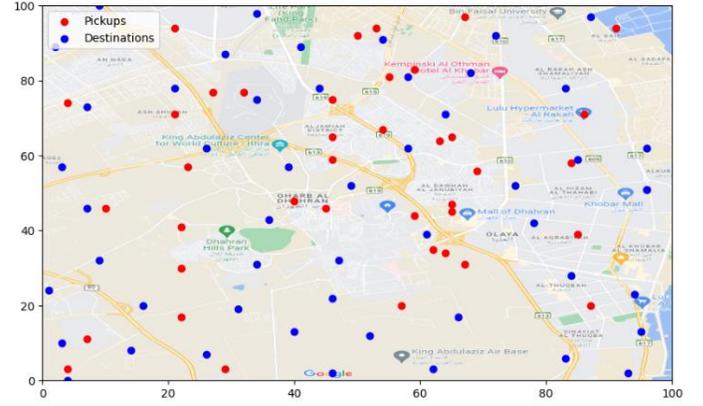

**Fig. 2.** Distribution of pickup and destination points of requests

TABLE I
COST PARAMETERS

| Parameter | Value (MU) |
|---|---|
| Revenue per request, $f_r$ | 60 |
| Committed courier wage per *period*, $\omega^c$ | 1 |
| Committed courier travel cost per unit distance, $v^c$ | 0.1 |
| Occasional courier payment per request, $\omega^o$ | 2 |
| Occasional courier travel cost per unit distance, $v^o$ | 0.1 |
| Lost request penalty cost, $\theta$ | 60 |
| Compensation for shift extension per *period*, $\rho$ | 3 |

TABLE II
INPUT VALUES OF TIME AND EXTENSION PARAMETERS

| Parameter | Value |
|---|---|
| The average courier speed, $\lambda$ | 100 |
| Number of *periods* for each shift extension, $\eta$ | 6 *periods* |
| Occasional courier being idle before leaving, $\psi$ | Poisson distributed with a mean equal to 1 period |
| Maximum number of notified couriers in a *period*, $\alpha_t$ | 3 couriers |
| Acceptance probability of shift extension for courier $m$, $pr_m$ | 0.7 |
| Period length, $\sigma$ | 6 min |

Specifically, the input to the first linear layer directly maps the state dimensions to a hidden layer of size 64. The output layer, on the other hand, maps the last hidden layer to the Q-values associated with each action in the action space. The DQN weights, denoted by $w$, are optimized using the Adam optimizer introduced by [18]. The optimization process involves adjusting the weights based on the chosen learning rate of 0.01. The parameters for the DQN algorithm are summarized in Table IV.

TABLE III
INPUT PARAMETERS FOR THE DQN ALGORITHM

| Parameter | Value |
|---|---|
| RM size | 100 |
| $|M_{sub}|$ | 64 |
| $N_{eposides}$ | 1000 |

| ε | 0.99999 |
| --- | --- |
| ξ | 0.99999 |
| *Min ε* | 0.01 |
| γ | 0.7 |
| ζ | 5 |
| Learning Rate | 0.01 |
| Optimizer | Adam Optimizer |

*B. Evaluating DQN training results*

This section assesses the algorithmic performance of the training phase, namely using loss metrics and reward outcomes. Figures 3 and 4 show the "episode reward" and "cumulative average episode reward", respectively, in the context of the agent's interaction with the environment. The episode reward is the total reward acquired by the agent in each episode, whereas the cumulative average episode reward represents the cumulative average of the episode rewards accumulated over all episodes up to the current one. This metric serves as a comprehensive measure of the agent's overall performance and tracks the trend of its reward accumulation throughout the learning process. Recall that an episode consists of 200 epochs and the time between two consecutive epochs is six minutes.

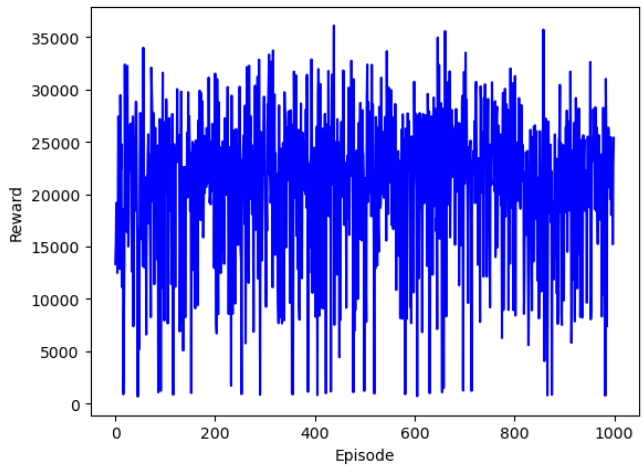

**Fig. 3.** Episode reward

As Figure 3 illustrates, granular fluctuations in episode reward lack interpretability; however, Figure 4 clearly depicts an increase in the cumulative average reward over successive episodes, which indicates that the agent is effectively leveraging experience within the environment to optimally approximate solutions that maximize reward acquisition over time. Similarly, Figures 5 and 6 illustrate episode loss and cumulative average episode loss over the training period of the reinforcement learning model. Figure 5 tracks loss on an episodic basis whereas Figure 6 presents the cumulative average of episode loss, revealing a downward trend. The decrease in the average loss shows that the algorithm is successfully learning from interactions with the environment to enhance its ability to predict optimal actions and maximize rewards. For all the experiments considered afterwards, we similarly verified the convergence of the trained models before reporting on the results.

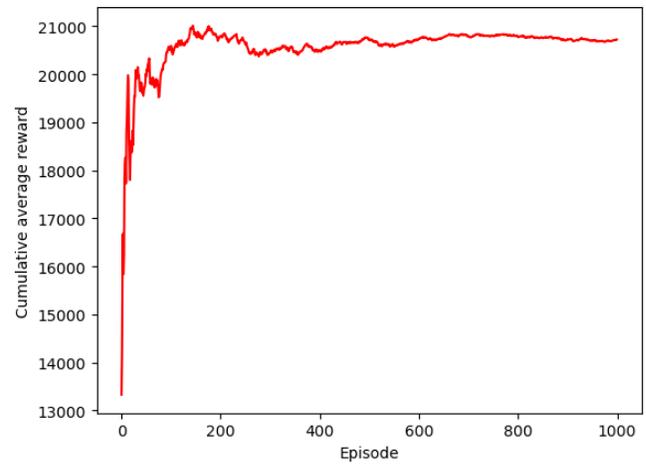

**Fig. 4.** Average episode reward

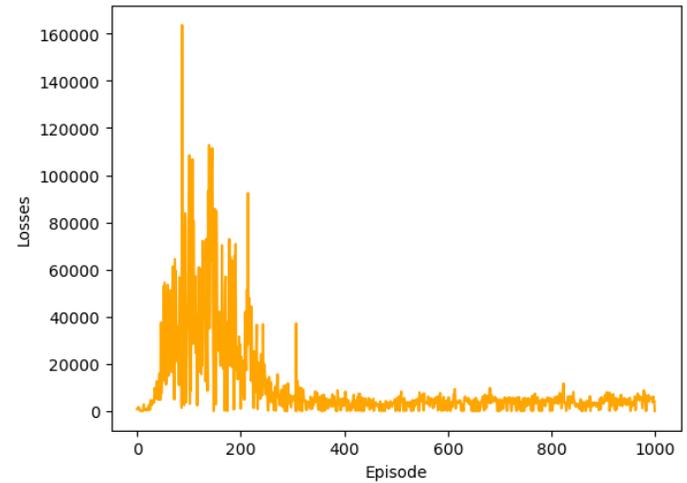

**Fig. 5.** Episode loss

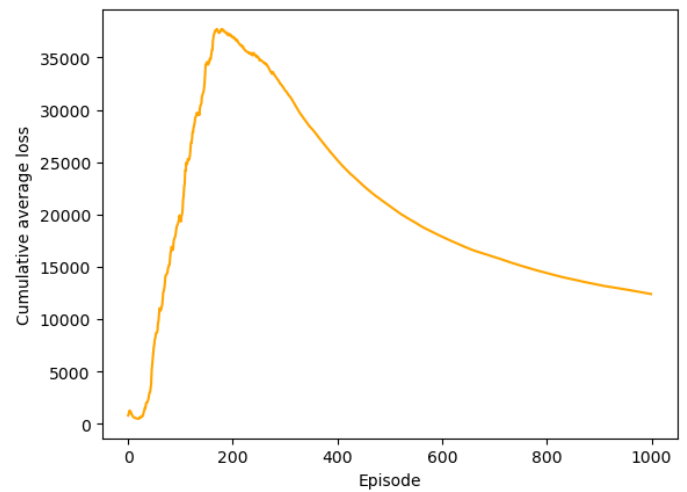

**Fig. 6.** Average episode loss

## C. Comparison of shift extension policy vs. no shift extension policy

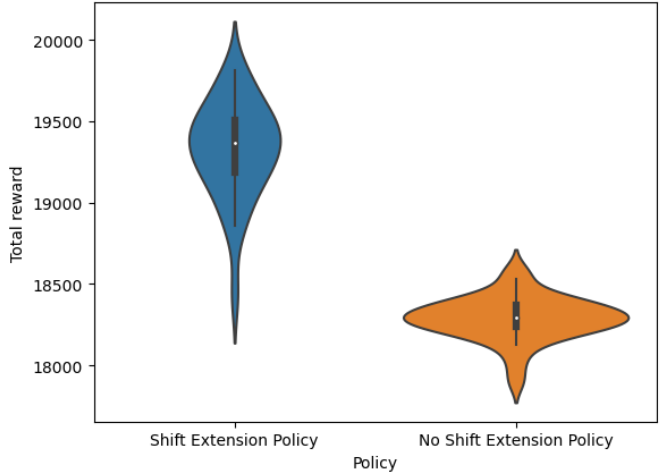

**Fig. 7.** Comparison of total reward (MU) of shift extension policy vs. no shift extension policy

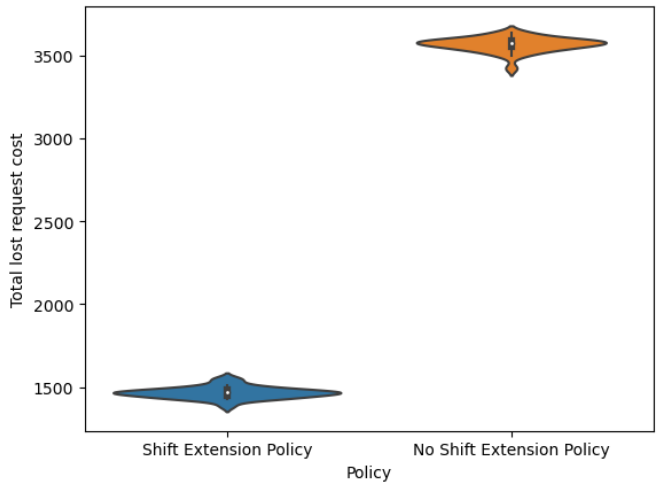

**Fig. 8.** Comparison of total lost request cost (MU) of shift extension policy vs. no shift extension policy

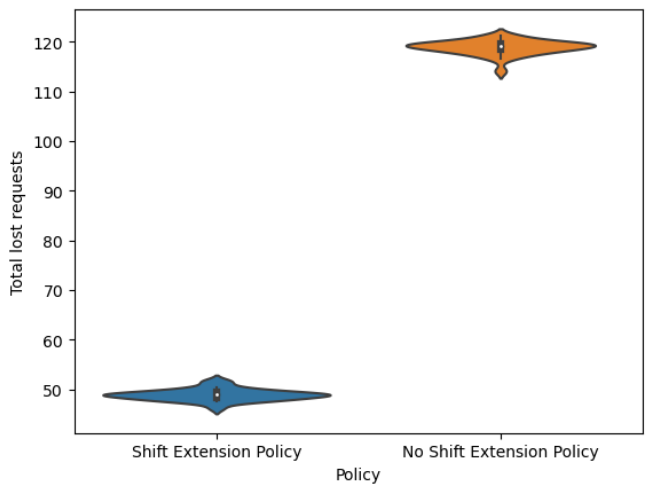

**Fig. 9.** Comparison of number of total lost requests of shift extension policy vs. no shift extension policy

In this section, we evaluate the effectiveness of the shift extension policy defined above through benchmarking with a baseline policy. We test our trained model in terms of the reward, the lost request cost, and the number of lost requests. We will compare the results of our model -*shifts extension policy, $\pi_s$*, with *no shift extension policy, $\pi_n$*, with respect to the aforementioned metrics. Thirty runs (each run is 1000 episodes) were conducted to gather sample averages for the metrics. Violin plots provide a visual representation of the distribution of the results of the 30 runs, offering insights into the density and central tendency of the depicted data. In this study, violin plots (Figures 7, 8, and 9) are created to visually represent the distributions of total reward, total lost request cost, and the total lost requests of the runs. The violin plot depicted in Figure 7 shows that the total reward of the $\pi_s$ policy has a higher median compared to the $\pi_n$ policy. In the comparison of total lost order costs metric, the violin plot presented in Figure 8 demonstrates that the model incorporating shift extension exhibited a reduced median compared to its no-extension counterpart. Figure 9 presents another crucial performance metric: the number of lost requests. Illustrated in the violin plot, the shift extension model exhibits a lower median in comparison to its no-shift-extension counterpart. Given that the platform received a total of 432 requests (based on the average of the request arrival distribution) throughout the day, the average number of lost requests in the shift extension and no shift extension models is 49 and 119, respectively. Consequently, the percentage of lost requests in the two models stands at 11% and 27%, respectively. Across all three metrics, the results emphasize the efficacy of the shift extension policy as a mechanism that can help increase total profit, and reduce the number of lost requests (increase the service level).

## D. Sensitivity Analysis

In order to study the flexibility and robustness of the proposed delivery system model, we conduct a sensitivity analysis study by systematically varying input model parameters and examining the effects on key output metrics. Specifically, in Sections V-D-1 and V-D-2, the arrival rate of requests and the arrival rate of occasional couriers per period will each be independently varied from their baseline values to observe their impact on the total extension cost. Additionally, in Section V-D-3, we vary the extension compensation to observe the impact on the service level (measured through the number of lost requests), which provides insight into the system's behavior under different pricing levels.

*1) The effect of high average number of requests vs. low average number of requests on extension cost*

Figure 10 shows violin plots comparing the total extension cost in three different scenarios: (1) *normal* arrival rate of new placed requests (Poisson distribution with the baseline mean of 2 requests per period used in previous sections), (2) *high* arrival rate of new placed requests (Poisson distribution with mean 3 requests per period), and (3) *low* arrival rate of new placed requests (Poisson distribution with mean 1 request per period), while keeping the offline schedule unchanged. The average total extension cost in the high, normal, and low scenarios were 205.06, 63.99, and 25.60, respectively.

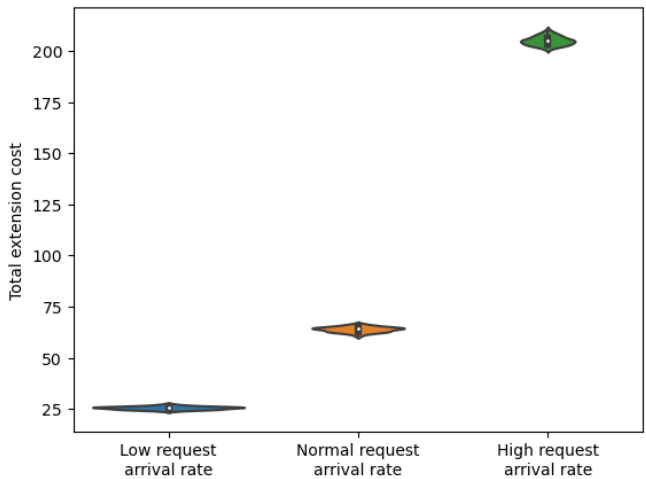

**Fig. 10.** Comparison of high request arrival rate vs. low request arrival rate on total extension costs

The results illustrate a nonlinear relationship between the customer request arrival rate and the average total extension cost. Specifically, the average total extension cost for the *high* scenario was approximately triple that of the *normal* scenario. On the other hand, the average total extension cost for the *normal* scenario was only about double that of the *low* scenario. This nonlinear response indicates that the number of extensions (and consequently, the total extension cost) is fairly sensitive to changes in the request arrival rate. It also demonstrates the model's ability to scale up or down its performance for a variety of request arrival rates.

The explanation behind these findings is that in the *low* scenario, where requests are less frequent, the committed couriers and occasional couriers can efficiently handle the workload without the need for many shift extensions. Consequently, the average total extension cost tends to decrease, leading to a lower median value. On the other hand, in the *high* scenario, where requests are more abundant, the existing couriers may not be sufficient to handle the workload. In such cases, shift extensions are necessary to optimize the reward. As a result, the average total extension costs increase, leading to a higher median value.

*2) The effect of high mean number of occasional couriers vs. low mean number of occasional couriers on extension cost*

Figure 11 presents violin plots comparing the total extension cost in three different scenarios for the occasional courier arrival rate: (1) *normal* (Poisson distribution with mean 1 courier per *period*), *high* (Poisson distribution with mean 2 couriers per *period*), and (3) *low* (Poisson distribution with mean 0.5 courier per *period*). The average total extension cost in the high, normal, and low scenarios are 41.37, 63.99, and 88.04, respectively. The results illustrate a linear relationship (with a slope almost equal one) between the arrival rate of occasional couriers and the average total extension cost. Specifically, the average total extension cost for the high occasional courier scenario was approximately 1.5 times that of the normal occasional scenario. Similarly, the mean total extension cost for the normal occasional scenario was about 1.5 times compared to the low occasional scenario.

We explain these findings by noting that in the case of a high number of occasional couriers, we are less likely to resort to shift extensions. This is because the priority is given to occasional couriers due to their lower cost compared to committed couriers. As a result, extension costs are lower in this scenario, leading to a lower median value. On the other hand, in the low scenario, occasional couriers are less abundant, and, therefore, the priority becomes to extend the shifts of committed couriers, resulting in higher average total extension costs.

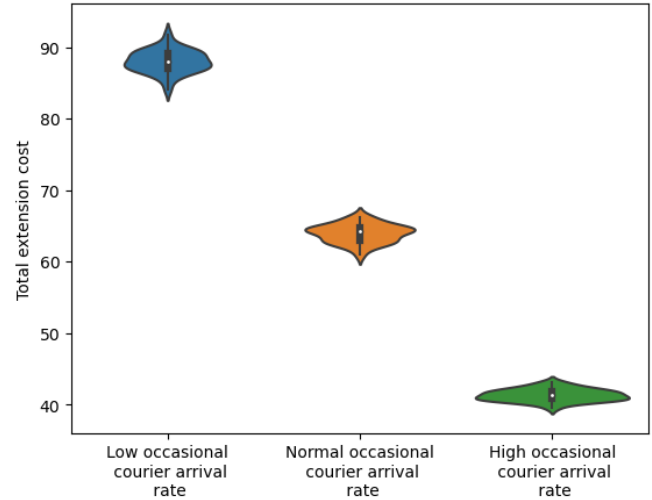

**Fig. 11.** Comparison of high occasional couriers arrival rate vs. low occasional couriers arrival rate on total extension costs.

*3) The effect of high shift extension compensation vs. low shift extension compensation on the service level*

Table IV summarizes the average number of extensions and the service level (measured through the average number of lost requests) for three scenarios based on compensation levels: (1) normal (the extension compensation aligns with the trained model), (2) high (compensation involves doubling the normal compensation), and (3) low (compensation is the half of the normal compensation). It is important to note that the acceptance probability for shift extensions was adjusted accordingly, as these parameters are interconnected. Specifically, as the extension compensation doubled, the probability of accepting the shift extension was increased from 0.7 to 0.9, and when the extension compensation is half the normal value, the probability of accepting the shift extension was decreased from 0.7 to 0.5. Table IV demonstrates that the *normal* scenario has the highest average number of extensions and lowest average lost requests compared to the *low* and *high* scenarios.

The reason for these findings is that when the extension compensation is low, although it is advantageous for the platform to send a higher number of notifications, the acceptance probability is low leading to an increase in the number of lost requests. Conversely, when the compensation is high, it becomes appealing for committed couriers to accept shift extensions. However, the platform sends notifications less frequently due to the associated costs, preferring to have lost requests rather than opting for shift extensions. As a result, the number of lost requests increases. These results highlight the importance of dynamic pricing of the shift extension compensation and how the DQN successfully learned these dynamics.

TABLE IV
COMPARISON OF HIGH SHIFT EXTENSION COMPENSATION VS. LOW SHIFT EXTENSION COMPENSATION ON THE SERVICE LEVEL (BEST VALUES ARE IN BOLD)

| | | Average lost requests | Average number of extensions |
|---|---|---|---|
| Scenario | High compensation | 74.0 | 18.0 |
| | Normal compensation | **49.0** | **21.3** |
| | Low compensation | 103.8 | 16.3 |

## VI. CONCLUSION

In conclusion, our study focuses on the dynamic adjustment of offline schedules through shift extensions for committed couriers in crowdsourced delivery platforms. We have developed a deep Q-network (DQN) learning framework to address the online scheduling problem of committed couriers, aiming to maximize platform profit by efficiently allocating requests and determining shift extensions. Benchmarking against a baseline policy that does not utilize shift extensions, our proposed shift extension policy demonstrates superior performance in terms of reward acquisition, reduced lost requests costs, and lost requests. The results showed that with shift extensions, the average number of lost requests was 49 out of 432 total requests received, a 11% loss rate, compared to 119 losses without extensions, a 27% rate. Therefore, allowing shift extensions significantly improved the service level by reducing lost requests by over half compared to not utilizing this capability.

Sensitivity analysis provided useful insights into how the proposed deep reinforcement learning model performs under different operational conditions. Variations in key parameters like requests arrival rate and occasional courier arrival rate impacted extension costs as expected. It also showed how the level of extension compensation influences the number of shift extensions and the number of lost requests. The results demonstrated a non-linear relationship between customer requests and extension costs. However, the relationship between occasional courier availability and costs was linear. In particular, the extension costs doubled when the arrival rate of occasional couriers in the normal condition was twice that of the low scenario. Furthermore, the costs doubled again when the arrival rate in the high scenario was twice that of the normal condition, which suggests a direct proportional scaling. However, when the arrival rates of requests decreased from normal to low (half the normal rate), the extension costs decreased by half. Conversely, when the arrival rates increased from normal to high (1.5 times the normal rate), the costs increased by threefold. Finally, when it comes to extension compensation, the normal scenario had the highest average number of shift extensions compared to the low (half the normal) and high (double the normal) scenarios. As a result, it had the fewest lost requests in comparison to the high or low compensation scenarios. These findings emphasize the significance of dynamically pricing the compensation for shift extensions and demonstrate the success of DQN in learning such dynamics.

Our findings further provide evidence that DRL is a promising technique for handling complex scheduling decisions in crowdsourced delivery faced with stochastic demands and resource availability. The shift extension formulation allowed the agent to learn dynamic policies that better adapt to changing conditions in real-time.

Going forward, there are a few potential avenues for further research. One option is applying alternative reinforcement learning algorithms, such as Proximal Policy Optimization (PPO), to the problem of online scheduling through courier shift extensions. This could provide a point of comparison to assess different algorithms' performance. Another direction to explore is building the schedule dynamically by adding and removing blocks in the shifts of committed couriers. Additionally, leveraging hierarchical reinforcement learning by jointly modeling scheduling, routing, and assignment decisions in an integrated manner. Taking a more holistic view that combines these different but related decisions could potentially yield even better results. Overall, there remain opportunities to build on this work by exploring various reinforcement learning techniques, and considering more sophisticated and integrated frameworks that tackle multiple decision layers such as Multiagent systems.

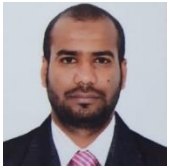
**Zead Saleh** is a PhD candidate in the Industrial and Systems Engineering Department at King Fahd University of Petroleum and Minerals (KFUPM), Saudi Arabia. He received his MS and BSc degree in Industrial Engineering from the King Fahd University of Petroleum and Minerals. He is a member of IISE institute. His research interests are in Logistics, Machine Learning, and Data Mining.

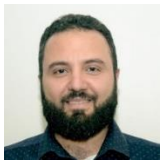
**Dr. Ahmad Al Hanbali** is an associate professor in the Department of Industrial and Systems Engineering at KFUPM. Before joining KFUPM, he worked as an assistant and associate professor in the Department of Industrial Engineering and Business Information Systems at the University of Twente, The Netherlands. Currently, he serves as the area editor of Systems Engineering in the Arabian Journal of Science and Engineering (Springer). He is(was) a full member of the Beta Research school, The Netherlands, the International Society of Inventory Research (ISIR), Production and Operations Management Society (POMS), IISE institute, and IEOM Society.

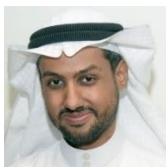
**Dr. Ahmad Baubaid** is an assistant professor at the Industrial and Systems Engineering Department, KFUPM. He received his PhD degree in Industrial Engineering from the Georgia Institute of Technology, Atlanta, USA in 2020. Prior to that, he received his MS degree in Operations Research from the Georgia Institute of Technology in 2016, and his BSc degree in Industrial Engineering from the King Fahd University of Petroleum and Minerals. His research interests are in transportation, logistics, and network design. He is a member of INFORMS and the INFORMS Transportation Science and Logistics Society (TSL).